\documentclass[11pt,a4paper]{article}
\usepackage[hyperref]{emnlp2020}
\usepackage{times}
\usepackage{latexsym}

\usepackage{graphicx}
\usepackage{multirow}
\usepackage{xcolor}
\usepackage{xcolor,colortbl}
\usepackage{microtype}

\usepackage{float}
\usepackage{booktabs}
\usepackage{ctable}
\usepackage{csquotes}
\usepackage{comment}
\usepackage{enumitem}

\aclfinalcopy 

\title{Misinformation Has High Perplexity}

\author{Nayeon Lee\thanks{$^*$ These two authors contributed equally.} , Yejin Bang$^*$, Andrea Madotto, Pascale Fung\\
  Center for Artificial Intelligence Research (CAiRE) \\
  Hong Kong University of Science and Technology \\
  {\tt [nyleeaa,yjbang,amadotto].connect.ust.hk, pascale@ece.ust.hk} }
\date{}

\begin{document}
\maketitle
\begin{abstract}
Debunking misinformation is an important and time-critical task as there could be adverse consequences when misinformation is not quashed promptly. However, the usual supervised approach to debunking via misinformation classification requires human-annotated data and is not suited to the fast time-frame of newly emerging events such as the COVID-19 outbreak. 
In this paper, we postulate that misinformation itself has higher perplexity compared to truthful statements, and propose to leverage the perplexity to debunk false claims in an unsupervised manner. First, we extract reliable evidence from scientific and news sources according to sentence similarity to the claims. Second, we prime a language model with the extracted evidence and finally evaluate the correctness of given claims based on the perplexity scores at debunking time. We construct two new COVID-19-related test sets, one is scientific, and another is political in content, and empirically verify that our system performs favorably compared to existing systems. We are releasing these datasets\footnote{https://github.com/HLTCHKUST/covid19-misinfo-data} publicly to encourage more research in debunking misinformation on COVID-19 and other topics.
\end{abstract}


\section{Introduction}
\label{section:introduction}
Debunking misinformation is a process of exposing the falseness of given claims based on relevant evidence. The failure to debunk misinformation in a timely manner can result in catastrophic consequences, as illustrated by the recent death of a man who tried to self-medicate with chloroquine phosphate to prevent  COVID-19~\cite{vigdor_2020}. Amid the COVID-19 pandemic and infodemic, the need for an automatic debunking system has never been more dire.
\begin{figure}[t]
    \small
    \centering
    \includegraphics[width=0.9\linewidth]{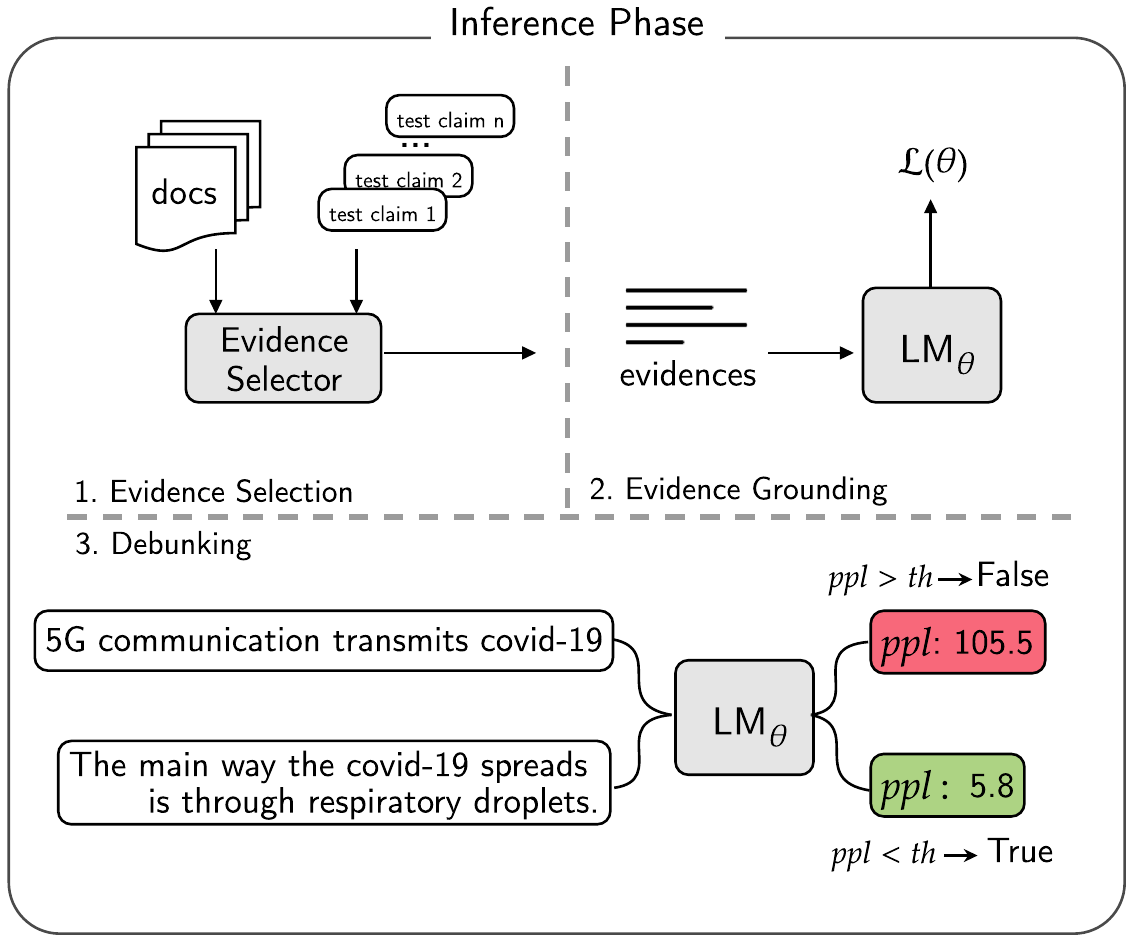}
    \caption{Proposed approach of using the language model (LM) as a debunker. We prime an LM with extracted evidence relevant to the whole set of claims, and then we compute the perplexities during the debunking stage.}
    \label{fig:ppl_illustration_for_pg1}
\end{figure}

A lack of data is the major obstacle for debunking misinformation related to newly emerged events like the COVID-19 pandemic. There are no labeled data available to adopt supervised deep-learning approaches \cite{Etzioni2008,wu2014toward,ciampaglia2015computational,popat2018declare, alhindi2018your}, and an insufficient amount of social- or meta-data to apply feature-based approaches \cite{long2017fake, karimi2018multi, kirilin2018exploiting}. To overcome this bottleneck, we propose an unsupervised approach of using a large language model (LM) as a debunker.  

Misinformation is a piece of text that contains false information regarding its \textit{subject}, resulting in a rare co-occurrence of the \textit{subject} and its neighboring words in a truthful corpus. When a language model primed with truthful knowledge is asked to reconstruct the missing words in a claim, such as ``5G communication transmits ... '', it would predict word ``information'' with the highest probability ($7\times10^{-1}$). On the other hand, it would predict the word "COVID-19" in a false claim such as ``5G communication transmits COVID-19'' with very low probability  (i.e., $2\times10^{-4}$). It follows that truthful statements would give low perplexity whereas false claims tend to have high perplexity, when scored by a truth-grounded language model. 
Since perplexity is a score for quantifying the likelihood of a given sentence based on previously encountered distribution, we propose \textit{a novel interpretation of perplexity as a degree of falseness}.

To further address the problem of data scarcity, we leverage the large pre-trained LM, such as GPT-2 \cite{radford2019language}, which are shown to be helpful in a low-resource setting by allowing the transfer learning of features learned from their huge training corpus~\cite{devlin2018bert,radford2018improving,liu2019roberta,lewis2019bart}. It is also illustrated the potential of LMs in learning useful knowledge without any explicit task-specific supervision to perform well on tasks such as question answering and summarization~\cite{radford2019language, petroni2019language}.

Moreover, it is crucial to ensure that the LM is populated with ``relevant and clean evidence" before assessing the claims, especially when these are related to newly emerging events. There are two main ways of obtaining evidence in fact-checking tasks. One way is to rely on evidence from the structured knowledge base such as Wikipedia and knowledge-graph~\cite{wu2014toward, ciampaglia2015computational, thorne2018fever, yoneda-etal-2018-ucl, nie2019combining}. Another approach is to obtain evidence directly from unstructured data online~\cite{Etzioni2008,magdy2010web}. However, the former approach faces a challenge in maintaining up-to-date knowledge, making it vulnerable to unprecedented outbreaks. On the other hand, the latter approach suffers from the credibility issue and the noise of the evidence. In our work, we attempt to combine the best of both worlds in the evidence selection step by extracting evidence from unstructured-data and ensuring quality by filtering noise.

The contribution of our work is threefold. First, we propose a novel dimension of using language model perplexity to debunk false claims, as illustrated in Figure \ref{fig:ppl_illustration_for_pg1}. It is not only data efficient but also achieves promising results comparable to supervised baseline models. We also carry out qualitative analysis to understand the optimal ways of exploiting perplexity as a debunker. Second, we present an additional evidence-filtering step to improve the evidence quality, which consequentially improves the overall debunking performance. Finally, we construct and release two new COVID-19-pandemic-related debunking test sets.

\begin{table*}[t]
\centering
\resizebox{0.90\linewidth}{!}{%
    \begin{tabular}{lclc}
    \toprule
    \textbf{False Claims} & \textbf{Perplexity} \\ \midrule
    Ordering or buying products shipped from overseas will make a person get COVID-19. & 556.2  \\
    Sunlight actually can kill the novel COVID-19. & 385.0 \\
    5G helps COVID-19 spread. & 178.2 \\
    Home remedies can cure or prevent the COVID-19. & 146.2 \\\midrule\midrule

    \textbf{True Claims} & \textbf{Perplexity} \\ \midrule
    The main way the COVID-19 spreads is through respiratory droplets. & 5.8 \\
    The most common symptoms of COVID-19 are fever, tiredness, and dry cough. & 6.0 \\
    The source of SARS-CoV-2, the coronavirus (CoV) causing COVID-19 is unknown. & 8.1 \\
     Currently, there is no vaccine and no specific antiviral medicine to prevent or treat COVID-19. & 8.4 \\ 
     \bottomrule
    \end{tabular}
 } %
\caption{Relationship between claims and perplexity. False claims have higher perplexity compared to True claims.}
\label{table:ppl_false_true_examples}
\end{table*}

\section{Motivation}
\paragraph{Language Modeling}
Given a sequence of tokens $X=\{x_0,\cdots,x_n\}$, language models are trained to compute the probability of the sequence $p(X)$. This probability is factorized as a product of conditional probability by applying the chain rules \cite{manning1999foundations,bengio2003neural}:

\begin{equation}
    p(X)=\displaystyle\prod_{i=1}^{n} p(x_{i} | x_{0}, \cdots, x_{i-1})
\end{equation}
In recent years, large transformer-based~\cite{vaswani2017attention} language models have been trained to minimize the negative log-likelihood over a large collection of documents. 

\paragraph{Leveraging Perplexity}
Perplexity, a commonly used metric for measuring the performance of an LM, is defined as the inverse of the probability of the test set normalized by the number of words:
\begin{equation}
    PPL(X) = \sqrt[n]{\displaystyle\prod_{i=1}^{n} \frac{1}{p(x_{i} |  x_{0}, \cdots, x_{i-1})}} 
\end{equation}
Another way of interpreting perplexity is the measure of the likelihood of a given test sentence in reference to the training corpus. Based on this intuition, we hypothesize the following:

\begin{displayquote}
\textit{``When a language model is primed with a collection of relevant evidence about given claims, the perplexity can serve as an indicator for falseness."}
\end{displayquote}

The rationale behind is that the extracted evidence sentences for a $\texttt{True}$ claim would share more similarities (e.g., common terms or synonyms) with its associated claim. This leads to $\texttt{True}$ claims to have lower perplexity while the $\texttt{False}$ claims remain having higher perplexity. 



\section{Methodology}
\label{section:approach}
\paragraph{Task Definition}
Debunking is the task of exposing the falseness of a given claim by extracting relevant evidence and verifying the claims upon it. 
More formally, given a claim $c$ with its corresponding source document $\mathcal{D}$, the task of debunking is to assigning a label from  $\{\texttt{True}, \texttt{False}\}$ by retrieving and utilizing a relevant set of evidence $\mathcal{E}=\{e_{1}, e_{2}, e_{3}\}$ from $\mathcal{D}$.

Our approach involves three steps in the inference phase: 1) Evidence selection to retrieve the most relevant evidence from $\mathcal{D}$.  2) Evidence grounding step to obtain our evidence-primed language model (LM Debunker). 3) Debunking step to obtain perplexity scores from the evidence-primed LM Debunker and debunking labels. 

\subsection{Evidence Selection}
\label{section:evidence_selection}
Given a claim $c$, our Evidence Selector retrieves the top-3 relevant evidence $\mathcal{E}=\{e_{1}, e_{2}, e_{3}\}$ in the following two steps.

\paragraph{Evidence Candidate Selection} 
Given the source documents $\mathcal{D}$, we select the top-10 most relevant evidence sentences for each claim. Depending on the domain of the claim and source documents, we rely on generic TF-IDF method to select the tuples of evidence candidates with their corresponding relevancy scores $\{{(e}_1, s_1),\cdots,({e}_{10}, s_{10})\}$. Note that any evidence extractor can be used here.


\paragraph{Evidence Filtering}  
After selecting the top candidate tuples for the claim $c$, we attempt to filter out the noise and unreliable evidence based on the following rulings:

1) When an evidence candidate is a quote from a low-credibility speaker such as an Internet meme\footnote{An idea, image, or video that is spread very quickly on the internet.} or social-media-post, we discard it (e.g., ``\textit{quote} according to a social media post.''). Note that this approach leverages the speaker information inherent in the extracted evidence; 2) If a speaker of the claim is available, any quote or statement by the speaker him/herself is inadmissible to the evidence candidate pool; 3) Any evidence identical to the given claim is considered to be ``invalid” evidence (i.e. direct quotation of true/false claim); 4) We filter out reciprocal questions, which only add noise but have no supporting or contradicting information to the claim from our observation. The examples of before and after this filtration is shown in the Appendix. 

The final top-3 evidence $\mathcal{E}$ is selected after the filtering based on the provided extractor score $s$. An example of a claim and its corresponding extracted evidence are shown in Table \ref{table:evidences_examples}.

\begin{table*}[t]
\centering
\resizebox{0.75\linewidth}{!}{
\begin{tabular}{lr}
\toprule
\textbf{Claim}: The main way the COVID-19 spreads is through respiratory droplets. & \textbf{Label}: \texttt{True    } \\ \midrule
\multicolumn{2}{l}{\begin{tabular}[c]{@{}l@{}}\textbf{Evidence 1}:  The main mode of COVID-19 transmission is via respiratory droplets, although the \\ potential of transmission by opportunistic airborne routes via aerosol-generating procedures \\ in health care facilities, and environmental factors, as in the case of Amoy Gardens, is known.\end{tabular}}\\
\midrule
\multicolumn{2}{l}{\begin{tabular}[c]{@{}l@{}}\textbf{Evidence 2}: The main way that influenza viruses are spread is from person to person via \\virus-laden respiratory droplets (particles with size ranging from 0.1 to 100 $\mu$m in diameter) that \\ are generated when infected persons cough or sneeze.\end{tabular}}\\\midrule
\multicolumn{2}{l}{\begin{tabular}[c]{@{}l@{}}\textbf{Evidence 3:} The respiratory droplets spread can occur only through direct person-to-person \\ contact or at a close distance.\end{tabular}}\\
\bottomrule
\end{tabular}
}
\caption{Illustration of evidence extracted using our Evidence Selector}
\label{table:evidences_examples}
\end{table*}

\subsection{Grounding LM with Evidence}
\label{section:lm_grounding}
For the purpose of priming the language model, all the extracted evidence for a batch of claims $\mathcal{C}=\{c_1,\cdots,c_k\}$ are aggregated as $E=\{e^1_{1},e^1_{2},e^1_{3},\cdots,e^k_{1},e^k_{2},e^k_{3}\}$. We obtain our evidence-grounded language model (LM Debunker) by minimizing the following loss $\mathcal{L}$: 
\begin{equation}
  \mathcal{L}(E) =  - \sum_{j=1}^{|E|} \sum_{i=1}^{n} \log p_{\theta}(x^j_i| x^j_{0}, \cdots, x^j_{i-1}),
  \label{eq:loss}
\end{equation}
where the $x^j_i$ denotes a tokens in the evidence $e^j$, and $\theta$ the parameters of the language model.
It is important to highlight that \textit{none of the debunking labels or claims are involved in this evidence grounding step} and that our proposed methodology is \textit{model agnostic}.

\subsection{Debunking with Perplexity}
\label{section:lm_debunking}
The last step is to obtain debunking labels based on the perplexity values from the LM Debunker. 
As shown in Table \ref{table:ppl_false_true_examples}, perplexity values reveal a pattern that aligns with our hypothesis regarding its association with falseness; the false claims have higher perplexity than the true claims (For more examples of perplexity values, refer to the Appendix).
Perplexity scores can be translated to debunking labels by comparing to a perplexity threshold $th$ that defines the $\texttt{False}$ boundary in the perplexity space. Any claim with a perplexity score higher than the threshold is classified as $\texttt{False}$, and vice versa for $\texttt{True}$.

The optimal method of selecting the hyper-parameter threshold $th$ is an open research question. From an application perspective, any value can serve as a threshold depending on the desired level of ``strictness'' towards false claims. We define ``strictness'' as the degree of precaution towards false negative error, which is the most undesirable form of error in debunking (refer to Section~\ref{section:analysis_and_discussion} for details). From an experimental analysis perspective, a small validation set could be leveraged for hyper-parameter tuning of the threshold ($th$). In this paper, since we have small test sets, we do k-fold cross-validation ($k=4$) to obtain the average performance reported in Section~\ref{section:results}.



\section{Dataset}
\label{section:dataset}
\subsection{COVID19 Related Test Sets}
\paragraph{Covid19-scientific} A new test set is constructed by collecting COVID-19-related myths and scientific truths labeled by reliable sources like MedicalNewsToday, Centers for Disease Control and Prevention (CDC), and World Health Organization (WHO). It consists of the most common scientific or medical myths about COVID-19, which must be debunked correctly to ensure the safety of the public (e.g., ``drinking a bleach solution will prevent you from getting the COVID-19.''). There are 142 claims (Table~\ref{table:data_statistics}) with labels obtained from the aforementioned reliable sources. According to WHO and CDC, some myths are unverifiable from current findings, and we assigned $\texttt{False}$ labels to them (e.g., ``The coronavirus will die off when temperatures rise in the spring.'').

\paragraph{Covid19-politifact} Another test set is constructed by crawling COVID-19-related claims fact-checked by journalists from a website called Politifact \footnote{https://www.politifact.com/}. Unlike the Covid19-scientific test set, it contains non-scientific and political claims such as ``For the coronavirus, the death rate in Texas, per capita of 29 million people, we're one of the lowest in the country''. Such political claims may not be life-and-death matters, but they still have the potential to bring negative sociopolitical effects. Originally, these claims are labeled into six classes \{pants-fire, false, barely-true, half-true, mostly-true, true\}, which represent the decreasing degree of fakeness. We use a binary setup for consistency with our setup for Covid19-scientific by assigning the first three classes as $\texttt{False}$ and the rest as $\texttt{True}$. For detailed data statistics, refer to Table \ref{table:data_statistics}.

\begin{table}[t]
\small
\centering
\resizebox{0.9\linewidth}{!}{
\begin{tabular}{cccc}
\toprule
Test sets & \begin{tabular}[c]{@{}c@{}}False \\ claims\end{tabular} & \begin{tabular}[c]{@{}c@{}}True \\ claims\end{tabular} & Total \\ \midrule
Covid19-scientific & 101 & 41 & 142 \\
Covid19-politifact & 263 & 77 & 340 \\ \bottomrule
\end{tabular}
}
\caption{COVID-19 Related Test Set Statistics}
\label{table:data_statistics}
\end{table}

\paragraph{Gold Source Documents} 
Different gold source document $\mathcal{D}$ are used depending on the domain of the test sets. 
For the Covid19-scientific, we use CORD-19 dataset\footnote{https://www.kaggle.com/allen-institute-for-ai/CORD-19-research-challenge}, a free open research resource for combating the COVID-19 pandemic. 
It is a resource of over 59,000 scholarly articles, including over 47,000 with full text, about COVID-19, SARS-CoV-2, and other related coronaviruses. 

For the Covid19-politifact, we leverage the resources of the Politifact website. When journalists verify the claims on Politifact, they provide pieces of text that contain: i) the snippets of relevant information from various sources, and ii) a paragraph of their justification for the verification decisions. We \textit{only} take the first part (i) to be our gold source documents, to avoid using explicit verdicts on test claims as evidence.

\begin{table*}[t]
\centering
\resizebox{0.9\linewidth}{!}{
\begin{tabular}{lccccccc}
\toprule
 & \multicolumn{3}{c}{Covid19-scientific} & \multicolumn{3}{c}{Covid19-politifact} \\ \cmidrule(lr){2-4} \cmidrule(lr){5-7}
\multicolumn{1}{l}{Models} & Accuracy & \begin{tabular}[c]{@{}c@{}}F1-Macro\end{tabular} & \begin{tabular}[c]{@{}c@{}}F1-Binary\end{tabular} & Accuracy & \begin{tabular}[c]{@{}c@{}}F1-Macro\end{tabular} & \begin{tabular}[c]{@{}c@{}}F1-Binary\end{tabular} \\ \midrule

Fever-HexaF & \cellcolor[HTML]{ECF4FF}64.8\% & \cellcolor[HTML]{ECF4FF}58.1\% & \cellcolor[HTML]{ECF4FF}74.8\% & \cellcolor[HTML]{ECF4FF}46.6\% & \cellcolor[HTML]{ECF4FF}37.9\% & \cellcolor[HTML]{ECF4FF}61.2\% \\

LiarPlusMeta & \cellcolor[HTML]{ECF4FF}42.3\% & \cellcolor[HTML]{ECF4FF}41.1\% & \cellcolor[HTML]{ECF4FF}32.8\% & \textbf{80.3\%} & 66.8\% & \textbf{86.5\%} \\

LiarPlus & \cellcolor[HTML]{ECF4FF}45.1\% & \cellcolor[HTML]{ECF4FF}44.8\% & \cellcolor[HTML]{ECF4FF}44.9\% & 54.4\% & 52.8\% & 61.5\% \\

LiarOurs & \cellcolor[HTML]{ECF4FF}61.5\% & \cellcolor[HTML]{ECF4FF}59.2\% & \cellcolor[HTML]{ECF4FF}82.8\% & 78.5\% & \textbf{67.7\%} & 86.4\% \\

\midrule
LM Debunker & \textbf{75.4\%} & \textbf{69.8\%} & \textbf{83.1\%} & 74.4\% & 58.8\% & 84.2\% \\ \bottomrule
\end{tabular}}
\caption{Result comparison of our LM Debunker and baselines on two COVID-19 related test sets. Blue highlights represent the models tested in out-of-distribution setting (i.e., train set and test set are from different distribution). Note that all accuracy scores are statistically significant ($\mathit{p}<0.05$). }
\label{table:result_table}
\end{table*}

\section{Experiments}
\label{section:experiment}
\subsection{Baseline Models}
Although unrelated to COVID-19 misinformation, there are notable state-of-the-art (SOTA) models and their associated datasets in fact-based misinformation field.

\paragraph{FEVER \cite{thorne2018fever}} Fact Extraction and Verification (FEVER) is a publicly released large-scale dataset generated by altering sentences extracted from Wikipedia to promote research in fact-checking systems. 
We use one of the winning systems from the FEVER workshop\footnote{https://fever.ai/2018/task.html. We use the 2nd team because we had problems running the 1st team's codebase. Note that the accuracy between 1st and 2nd is very minimal} as our first baseline model.

\begin{itemize}
    \item \textbf{Fever-HexaF}~\cite{yoneda2018ucl} Given a claim and a set of evidence, it leverages a natural language inference model to get entailment scores between claim and each evidence ($s_{e1}, s_{e2}, ..., s_{en}$), and obtains the final prediction label by aggregating the entailment scores using a Multi-Layer Perceptron (MLP).
\end{itemize}

\paragraph{LIAR-Politifact \cite{wang2017liar}} 
LIAR is a publicly available dataset collected from the Politifact website, which consists of 12.8k claims. The label setup is the same as our Covid19-politifact test set, and the data characteristics are very similar, but LIAR does not contain any claims related to COVID-19.

We also report three strong BERT-based~\cite{devlin2018bert} baseline models trained on LIAR data:
\begin{itemize}
    \item \textbf{LiarPlusMeta}: Our BERT-large-based replication of SOTA paper from~\citeauthor{alhindi2018your}. It uses meta-information and ``justification,'' human-written reasoning for verification decision in Politifact article, as evidence for the claim. Our replication is a more robust baseline, outperforming the reported SOTA accuracy by absolute $9\%$ (refer to Appendix for detailed result table).
    
    \item \textbf{LiarPlus}: Similar to LiarPlusMeta model, but without meta-information. Our replication also outperforms the SOTA by absolute $8\%$ in accuracy. This baseline is important because the focus of this paper is to explore the debunking ability in a data-scarce setting, where meta-information may not exist. 
    
    \item \textbf{LiarOurs}: Another BERT-large model fine-tuned on LIAR-Politifact claims with evidence from our Evidence Selector, instead of using human-written ``justification."
\end{itemize}

\subsection{Experiment Settings}
\label{section:experiment_settings}
\paragraph{Out-of-distribution Setting} For the Covid19-scientific test set, all models are tested in the out-of-distribution (OOD) setting because the test set is from different distribution compared to all the train sets used in baseline models; Fever-HexaF model is trained on FEVER dataset, all other Liar baseline models are trained on LIAR-Politifact dataset. 
For the Covid19-politifact test set, Fever-HexaF model is again tested in OOD setting. However, all the Liar-models are not because both their train sets and the Covid19-politifact test set are from a similar distribution (Politifact). We use blue highlights in the Table~\ref{table:result_table} to indicate models tested in OOD settings.

\paragraph{Evidence Input for Testing} Recalling the task definition explained in Section \ref{section:approach}, we test a claim with its relevant evidence. To make fair comparisons among all baseline models and our LM Denbunker, we use the same evidence extracted in the Evidence Selector step while evaluating the models on the COVID-19-related test sets. 

\paragraph{Language Model Setting}
For our experiments, GPT-2 \cite{Wolf2019HuggingFacesTS} model is selected as our base language model to build LM Debunker. We use the pre-trained GPT-2 model (base), with 117 million parameters. Since the COVID-19 pandemic is a recent event, it is guaranteed that the GPT-2 has not seen any COVID-19 related information during its pre-training. Thus, very clean and unbiased language model to test our hypothesis. 

\begin{table*}[t]
\centering
\resizebox{0.9\linewidth}{!}{
\begin{tabular}{lcccccc}
\toprule

\textbf{Covid19-politifact} 
& \multicolumn{3}{c}{LiarPlusMeta} &  \multicolumn{3}{c}{LiarOurs} \\\cmidrule(lr){1-1}\cmidrule(lr){2-4}
\cmidrule(lr){5-7}
\multicolumn{1}{l}{Train Size} & Accuracy  & F1-Macro & F1-Binary &  Accuracy  & F1-Macro & F1-Binary \\  \midrule
0 & N/A & N/A & N/A &  N/A & N/A & N/A \\
10 & 22.6\% & 18.5\% & 0.0\% &  22.6\% & 18.5\% & 0.0\% \\
100 & 22.6\% & 18.5\% & 0.0\% &  22.6\% & 18.5\% & 0.0\% \\
500 & 73.5\% & \cellcolor[HTML]{EFEFEF}65.2\% & 82.2\% & 64.4\% & \cellcolor[HTML]{EFEFEF} 59.4\% & 73.6\% \\
1000 & 72.4\% & 63.4\% & 81.5\% & 70.9\% & 64.2\% & 79.7\% \\
10000 (All) & \cellcolor[HTML]{EFEFEF}80.3\% & 66.8\% & \cellcolor[HTML]{EFEFEF}86.5\% & \cellcolor[HTML]{EFEFEF}78.5\% & 67.8\% & \cellcolor[HTML]{EFEFEF}86.4\% \\ \midrule
LM Debunker  & 74.4\% & 58.8\% & 84.2\% & 74.4\% &  58.8\% & 84.2\%\\
\bottomrule
\end{tabular}
}
\caption{Performance comparison between our LM Debunker and two baseline models trained on varying train set sizes. LiarPlusMeta and LiarOurs have shown the best performance on Covid-politifact test set in accuracy and F1-Macro, respectively. Gray highlights represent the first scores that surpass the LM Debunker scores.}

\label{table:result_train_size_change}
\end{table*}

\subsection{Experiment Details} 
We evaluate the performance LM Debunker by comparing it to other baselines on two commonly used metrics: accuracy and F1-Macro score. Since identifying $\texttt{False}$ claims is important in debunking, we also report F1 of $\texttt{False}$ class (F1-Binary).

Recall that we report \textit{average} results obtained k-fold cross-validation. The thresholds used in each fold are $\{18,19,17,22\}$ for Covid-politifact and $\{15,24,17,20\}$ for Covid-scientific\footnote{Note that we use k-fold cross-validation to obtain the average performance, not the average optimal threshold.}.

For the evidence grounding step, a learning rate of 5e-5 was adopted, and different epoch sizes were explored $\{1,2,3,5,10,20\}$. We reported the choice with the highest performance in both accuracy and F1-Macro. Each trial was run on Nvidia GTX 1080 Ti, taking 39 seconds per epoch for Covid-scientific and 113 seconds per epoch for Covid-politifact.

\section{Experimental Results}
\label{section:results}

\subsection{Performance of LM Debunker}
From Table \ref{table:result_table}, we can observe that our unsupervised LM debunker portrays notable strength in the out-of-distribution setting (highlighted with blue) over other supervised baselines. For the  Covid19-scientific test set, it achieved state-of-the-art results across all metrics and marginally improved in accuracy and F1-binary by an absolute $13.9\%\sim30.3\%$ and $10\%\sim28.7\%$ respectively. Considering the severe consequences Covid19-scientific myths could potentially bring, this result is valuable. 

For the Covid19-politifact test set, our LM debunker also outperformed Fever-HexaF model and LiarPlus with a significant margin, but it underperformed the LiarOurs model and the LiarPlusMeta model. Nonetheless, it is still encouraging considering the fact that these two models were trained with task-specific supervision on Politifact dataset (LIAR-Politifact), which is similar to the Covid19-politifact test set.

The results of LiarPlus and LiarPlusMeta clearly show the incongruity of the meta-based approach for cases in which meta-information is not guaranteed. LiarPlusMeta struggled to debunk the claims from the Covid19-scientific test set in contrast to achieving SOTA in Covid19-politifact test set. This is because the absence of meta-information for Covid19-scientific test set hindered LiarPlusMeta from performing to its maximum capacity. Going further, the fact that LiarPlusMeta performed even worse than LiarPlus emphasizes the difficulty faced by meta-based models in the absence of meta-information.



\begin{figure}[t]
    \small
    \centering
    \includegraphics[width=0.4\textwidth]{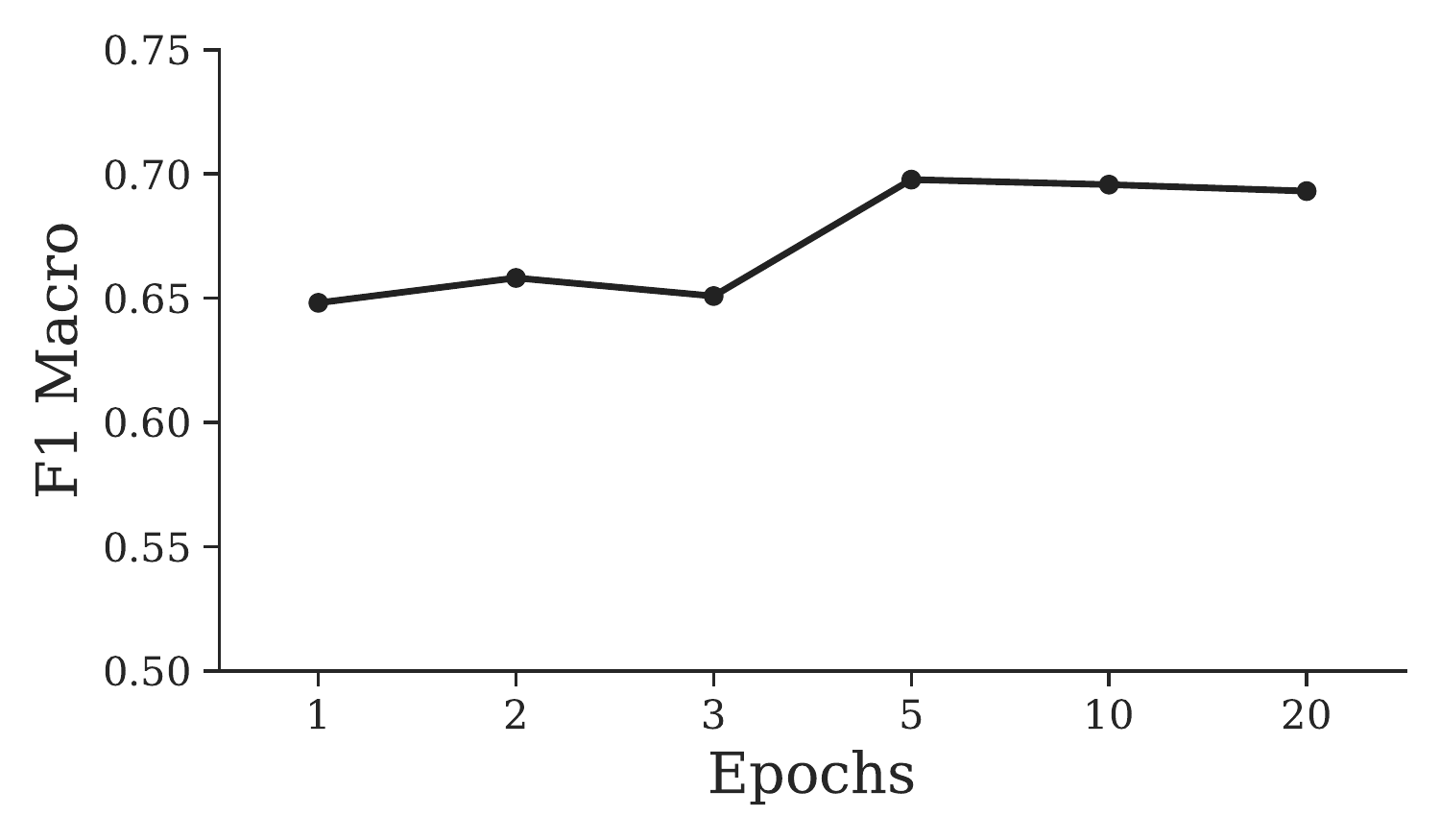}
    \caption{Trend of highest F1-Macro score over different numbers of epochs for evidence grounding}
    \label{fig:epoch-performance}
\end{figure}

FEVER dataset and many FEVER-trained systems successfully showcased the advancement of systematic fact-checking. Nevertheless, Fever-HexaF model exhibited rather low performance on COVID-19 test sets ($10\% \sim 30\%$ behind LM-debunker in accuracy). One possible justification is the way how FEVER data was constructed. FEVER claims were generated by altering sentences from Wikipedia (e.g., ``Hearts is a musical composition by Minogue", label: SUPPORTS). It makes the nature of FEVER-claims have a discrepancy with the naturally occurring myths and false claims flooding the internet. This implies that FEVER might not be the most suitable dataset for training non-wikipedia-based fact-checkers.

\subsection{Data Efficiency Comparison}
In Table \ref{table:result_train_size_change}, we report the performance of LiarOurs and LiarPlusMeta classifiers trained on randomly sampled train sets of differing sizes \{10, 100, 500, 1000, 10000\}. 

As shown by the gray highlights in Table \ref{table:result_train_size_change}, both classifiers overtake our debunker in F1-Macro score with $500$ labeled training data, but they require $10000$ to outperform on the rest of evaluation metric. 
Considering the scarcity of labeled misinformation data for newly emerged events, a data-efficient debunker is extremely meaningful.

\section{Analysis and Discussion}
\label{section:analysis_and_discussion}

\subsection{LM Debunker Behavior Analysis}
\paragraph{Number of Epoch for Evidence Grounding}

The relationship between the number of epoch in the evidence grounding step and the debunking performance is explored. The best performance is obtained with epoch=5, as shown in Figure ~\ref{fig:epoch-performance}. We believe this is because a low number of epochs does not allow enough updates to encode the content of evidence into the language model sufficiently. On the contrary, a higher number of epochs over-fit the language model to the given evidence and harms the generalizability of the language model. 

\begin{figure}[t]
    \small
    \centering
    \includegraphics[width=0.4\textwidth]{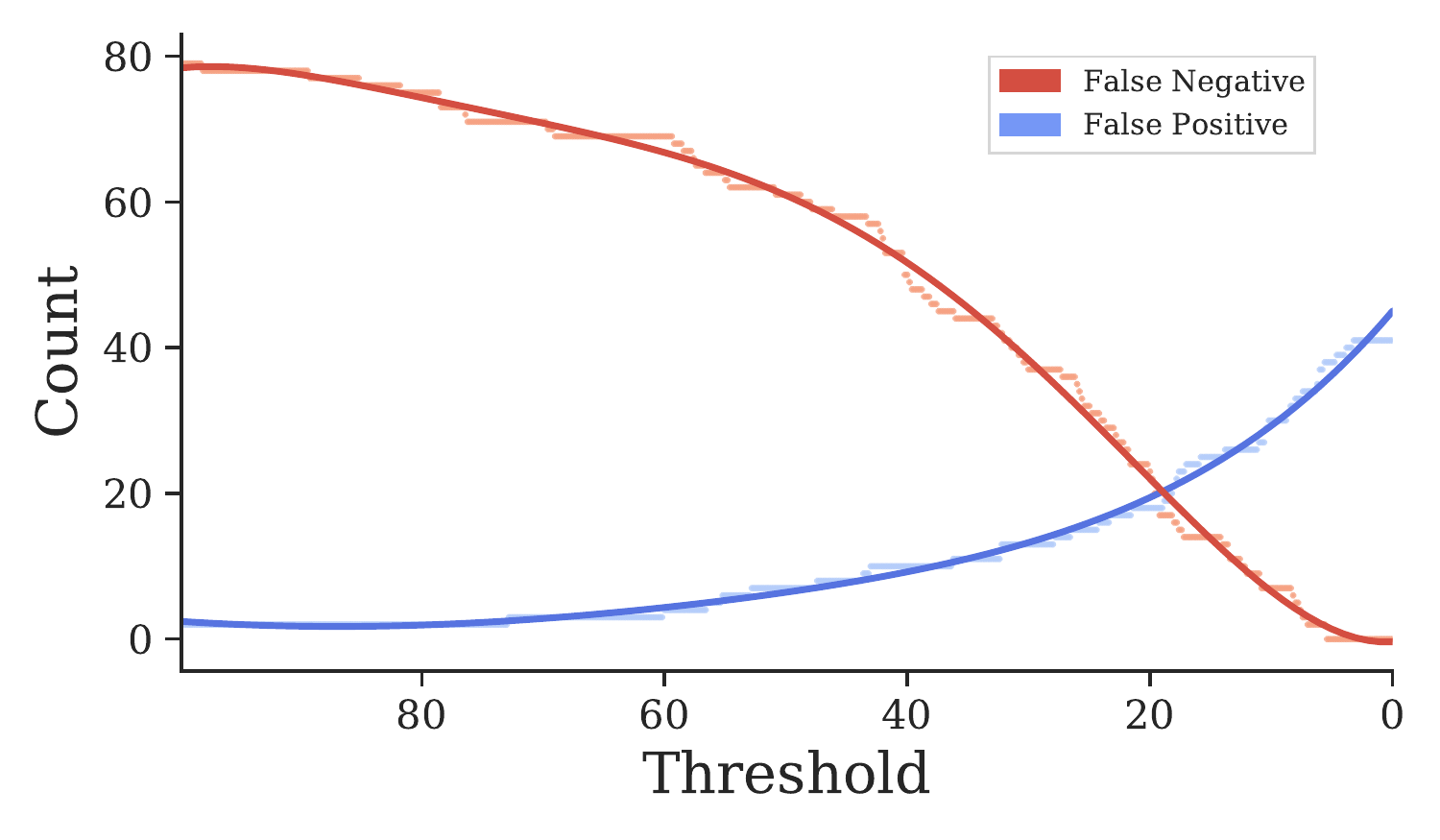}
    \caption{Trend of false negative and false positive counts over varying threshold.}
    \label{fig:threshold-performance}
\end{figure}

\paragraph{Threshold Perplexity Selection} 
As aforesaid, the threshold $th$ is controllable to reflect the desired ``strictness'' of the debunking system. Figure \ref{fig:threshold-performance} depicts that decreasing the threshold helps to reduce the FN errors, which is the most dangerous form of error. Such controllability over strictness would be beneficial to the real-world applications, where the level of ``strictness'' matters greatly depending on the purpose of the applications. 

Meanwhile, FN reduction comes with a trade-off of increased false positive errors (FP). For a more balanced debunker, an alternative threshold choice could be the intersection point of FN and FP frequencies.

\begin{table}[t]
\small
\resizebox{\linewidth}{!}{
\begin{tabular}{lcccc}
\toprule
 & \multicolumn{2}{c}{Covid19-scientific} & \multicolumn{2}{c}{Covid19-politifact} \\ \cmidrule(lr){2-3} \cmidrule(lr){4-5} 
 & Acc. & F1-Macro & Acc. & F1-Macro \\ \midrule
Before & 74.6\% & 56.3\% & \textbf{75.0\%} & 50.6\% \\
After & \textbf{75.4\%} & \textbf{69.8\%} & 74.4\% & \textbf{58.9\%} \\ \bottomrule
\end{tabular}
}
\caption{Performance comparison between the ``before'' and ``after'' filtering steps in Evidence Selector}

\label{ref:evidence_beforeandafter}
\end{table}

\subsection{Evidence Analysis}
Our ablation study of the evidence filtering and cleaning steps (Table \ref{ref:evidence_beforeandafter}) shows that improved evidence quality brings big gains in F1-Macro scores ($13.5\%$ and $8.3\%$) with only $1\%$ loss in accuracy.

Moreover, comparing the performances of LM Debunker on each of the two test sets, Covid19-scientific scores surpass Covid19-politifact scores, especially in F1-Macro, by $11.1\%$. 
This is due to the disparate natures of the gold source documents used in evidence selection; the Covid19-scientific claims obtain evidence from \textit{scholarly articles}, whereas Covid19-politifact claims extract evidence from \textit{news articles and other unverified internet sources}. Consequently, this resulted in a different quality of extracted evidence.
Therefore, an important insight would be that evidence quality is crucial to our approach, and additional performance gain would be fostered from further improvement in evidence quality.

\subsection{Error analysis and Future Work}  
We identified areas for improvement in future work through qualitative analysis of wrongly-predicted samples from the LM debunker. First, since perplexity originally serves as a measure of sentence likelihood, when a true claim has an abnormal sentence structure, our LM deunker makes a mistake by assigning high perplexity. For example, a true claim ``So Oscar Helth, the company tapped by Trump to profit from covid tests, is a Kushner company. Imagine that, profits over national safety'' has extremely high perplexity. One interesting future direction would be to explore a way of disentangling ``perplexity as a sentence quality measure'' from ``perplexity as a falseness indicator''.

Second, our LM debunker makes mistakes when selected evidence is refuting the \texttt{False} claim by simply negating the content of the paraphrased claim. For instance, 
for a false claim  ``Taking ibuprofen worsens symptoms of COVID-19,'' the top most relevant evidence from the scholarly articles is ``there is no current evidence indicating that ibuprofen worsens the clinical course of COVID-19." Another future direction would be to learn a better way of assigning additional weight/emphasis on special linguistic features such as negation.  

\section{Related Work}
\label{section:related_work}

\subsection{Misinformation}
Previous approaches~\cite{long2017fake, karimi2018multi, kirilin2018exploiting, shu2018understanding, monti2019fake} show that using meta-information (e.g.  credibility score of the speaker) with text input helps improve the performance of misinformation detection. Considering the availability of meta-information is not always guaranteed, building a model independent from it is crucial to detect misinformation. There exist works with fact-based approaches, using evidence from external sources for assessing the truthfulness of information \cite{Etzioni2008,wu2014toward,ciampaglia2015computational,popat2018declare, alhindi2018your, baly2018integrating,lee2018improving,hanselowski2018ukp}. These approaches based on the logic of ``the information is correct if evidence from credible sources or a group of online sources is supporting it." Furthermore, some works focus on reasoning and evidence selecting ability by restricting the scope of facts to those from Wikipedia~\cite{thorne2018fever,nie2019combining,yoneda-etal-2018-ucl}

\subsection{Language Model Applications}
lead to significant advancements in wide variety of NLP tasks, including question-answering, commonsense reasoning, and semantic relatedness \cite{devlin2018bert,radford2019language,peters2018deep,radford2018improving}. 
These models are typically trained on documents mined from Wikipedia (among other websites). 
Recently, a number of works have found that LMs store a surprising amount of world knowledge, focusing particularly on 
the task of open-domain question answering \cite{petroni2019language,roberts2020much}. Going further, \citeauthor{guu2020realm, roberts2020much} show that task specific fine-tuning of LM can achieve impressive results, proving the power of LMs. In this paper, we explore to confirm if large pre-trained LM can also be helpful in the field of debunking.

\section{Conclusion}
\label{section:conclusion}
In this paper, we show that misinformation has high perplexity from the language model primed with relevant evidence. By proposing the new application of perplexity, we build an unsupervised debunker that shows promising results, especially in the absence of labeled data. Moreover, we emphasize the importance of evidence quality in our methodology by showing the improvement in the final performance with the addition of a filtering step in the evidence selection. We are also releasing two new COVID-19 related test sets publicly to promote transparency and prevent the spread of misinformation. Based on this successful leverage of language model perplexity for debunking, we hope to foster more research in this new direction. 


\section*{Acknowledgements}
We would like to thank Madian Khabsa for the helpful discussion and inspiration.

\bibliography{emnlp2020}
\bibliographystyle{acl_natbib}

\clearpage
\clearpage
\begin{figure*}[t]
    \centering
    \includegraphics[width=\textwidth]{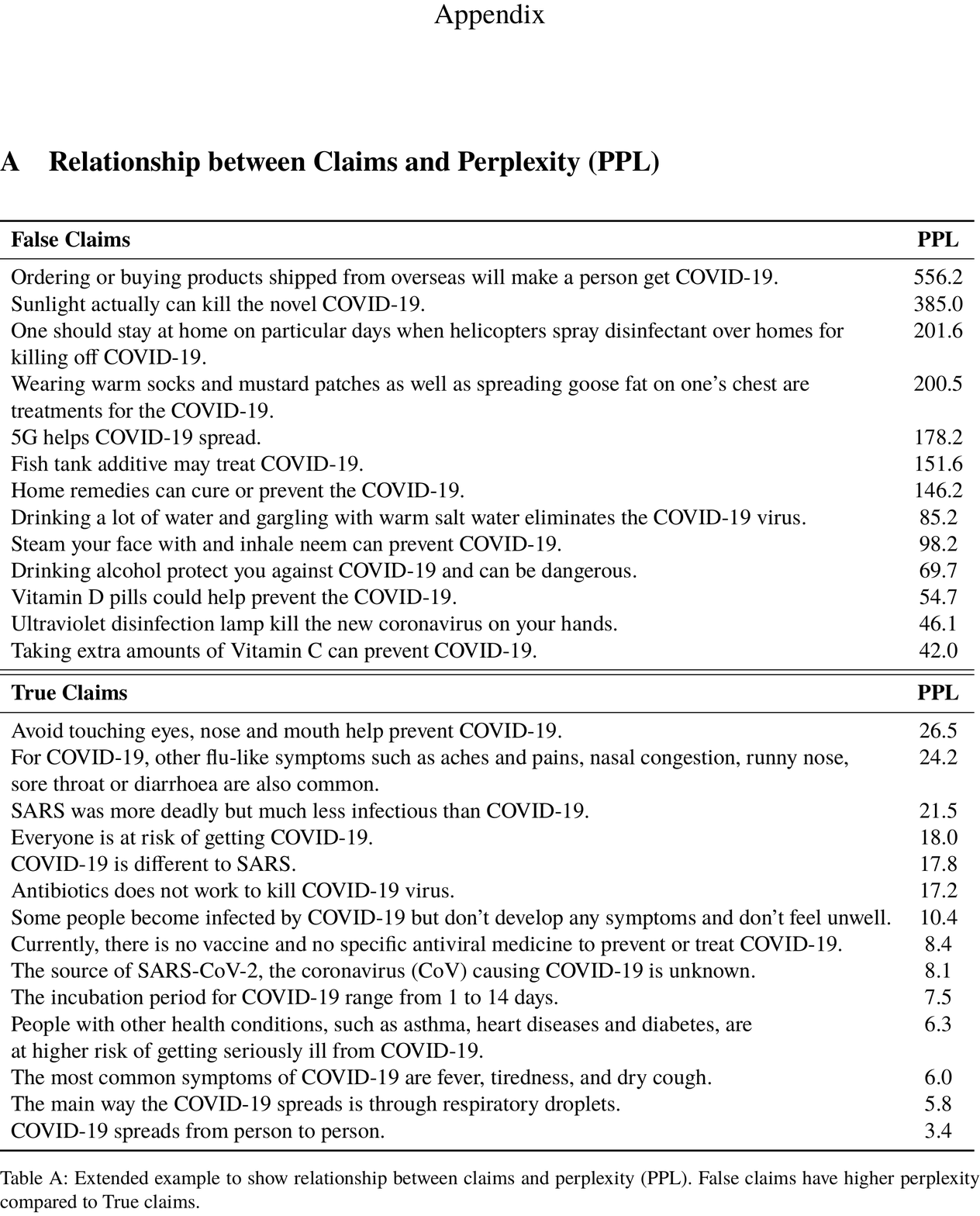}
\end{figure*}
\begin{figure*}[t]
    \centering
    \includegraphics[width=\textwidth]{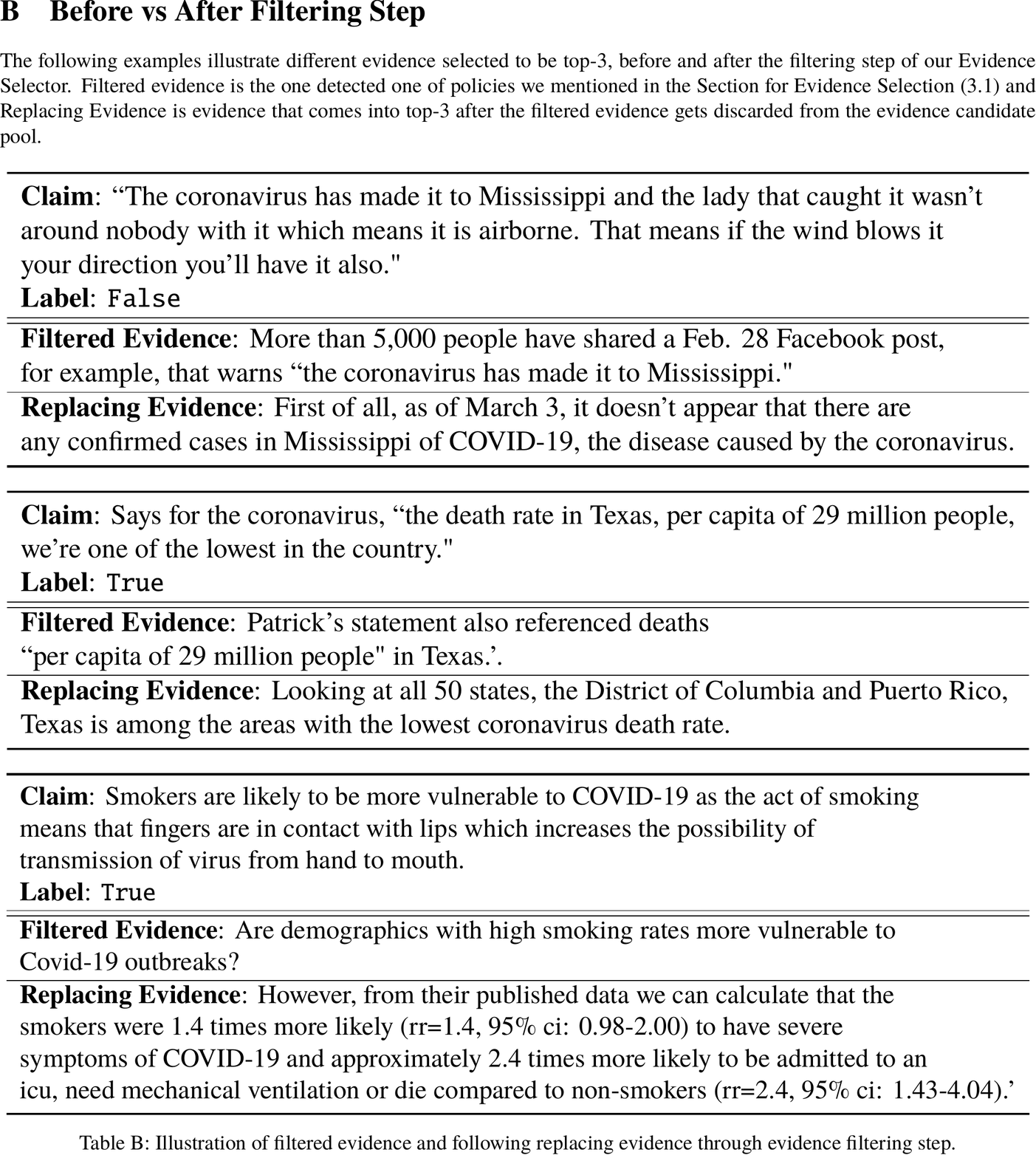}
\end{figure*}
\begin{figure*}[t]
    \centering
    \includegraphics[width=\textwidth]{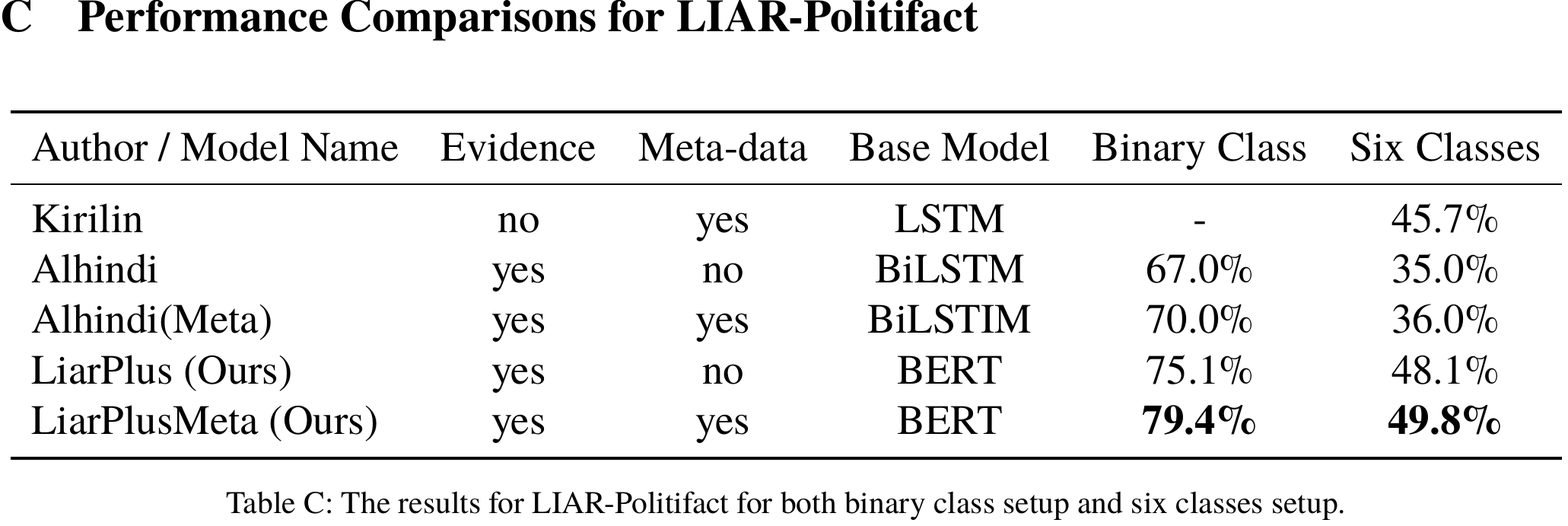}
\end{figure*}

\end{document}